\documentclass[sigconf,pdftex]{acmart}

\setcopyright{none}
\acmYear{2021}
\copyrightyear{The DOI of the original article is as follows}
\acmConference[This is the author version of the paper accepted \\ for GECCO '21]{2021 Genetic and Evolutionary Computation Conference}{July 10--14, 2021}{Lille, France}
\acmPrice{}
\acmISBN{}
\acmDOI{10.1145/3449639.3459290}

\usepackage{amsmath,amsfonts,mathtools}
\usepackage{balance}
\usepackage{bm}
\usepackage[]{graphicx}
\usepackage{textcomp}
\usepackage[]{xcolor}
\usepackage{algorithm,algorithmic}
\usepackage{cleveref}
\usepackage{subcaption}
\usepackage{booktabs}
\usepackage{seqsplit}

\begin{document}
\title{Level Generation for Angry Birds with Sequential VAE and Latent Variable Evolution}

\author{Takumi Tanabe}
\affiliation{%
\institution{University of Tsukuba \& RIKEN AIP}
}
\email{tanabe@bbo.cs.tsukuba.ac.jp}

\author{Kazuto Fukuchi}
\affiliation{%
\institution{University of Tsukuba \& RIKEN AIP}
}
\email{fukuchi@cs.tsukuba.ac.jp}

\author{Jun Sakuma}
\affiliation{%
\institution{University of Tsukuba \& RIKEN AIP}
}
\email{jun@cs.tsukuba.ac.jp}

\author{Youhei Akimoto}
\affiliation{%
\institution{University of Tsukuba \& RIKEN AIP}
}
\email{akimoto@cs.tsukuba.ac.jp}

\renewcommand{\shortauthors}{T. Tanabe et al.}

\begin{abstract}
  Video game level generation based on machine learning (ML), in particular, deep generative models, has attracted attention as a technique to automate level generation. However, applications of existing ML-based level generations are mostly limited to tile-based level representation.
  When ML techniques are applied to game domains with non-tile-based level representation, such as \emph{Angry Birds}, where objects in a level are specified by real-valued parameters, ML often fails to generate playable levels.
  In this study, we develop a deep-generative-model-based level generation for the game domain of \emph{Angry Birds}. To overcome these drawbacks, we propose a sequential encoding of a level and process it as text data, whereas existing approaches employ a tile-based encoding and process it as an image. Experiments show that the proposed level generator drastically improves the stability and diversity of generated levels compared with existing approaches. We apply latent variable evolution with the proposed generator to control the feature of a generated level computed through an AI agent's play, while keeping the level stable and natural.
\end{abstract}

%
%



\maketitle

\providecommand{\maxx}{\texttt{MAX\_COL}}
\providecommand{\maxy}{\texttt{MAX\_ROW}}
\providecommand{\groundy}{\texttt{GROUND\_Y}}
\providecommand{\platformy}{$y_\text{platform}$}
\providecommand{\eps}{\texttt{EPS}}


\section{Introduction}\label{sec:intro}
Procedural content generation (PCG) automatically generates the content of games using an algorithm.
The design of game content requires a large amount of knowledge and experience regarding the game.
Therefore, manually creating new game content is an expensive task.
Thus, PCG is a promising approach to minimize costs and generate new game content with different features, such as the difficulty of the game.
PCG has been recently applied to various games for different purposes \cite{hendrikx2013procedural}.
This study focuses on the automatic generation of game levels as a sub-category of PCGs.

Machine-learning-based PCG (PCGML) is a recent approach that has gathered research attention \cite{summerville2018procedural,liu2020deep}.
This approach generates video game content using machine learning (ML) models trained on existing game content as the dataset.
Human-designed game content satisfy implicit constraints such as whether a level is \seqsplit{playable} and has aesthetics.
ML models learn such implicit constraints from a dataset without implementing them manually.
Therefore, if the game content for a game is made available to the public, anyone can use it to build a game content generator without knowledge and experience regarding the game.

PCGML includes approaches that utilize n-gram \cite{dahlskog2014linear} with a graphical probability model \cite{guzdial2016game}, generative adversarial networks (GAN) \cite{volz2018evolving,torrado2019bootstrapping,giacomello2019searching}, and variational autoencoders (VAEs) \cite{thakkar2019autoencoder}.
Volz et al.~\cite{volz2018evolving} applied the Wasserstein GAN (WGAN) to generate levels for \emph{Super Mario Bros} (SMB).
Torrado et al.~\cite{torrado2019bootstrapping} successfully generated playable levels with a small dataset using conditional embedding self-attention GAN and the bootstrapping method.
Thakkar et al.~\cite{thakkar2019autoencoder} compared the autoencoder and VAE on game-level generation and demonstrated that the VAE can generate more complex levels.

An advantage of PCGML using a deep generative models (DGMs), such as GAN or VAE, is the possibility of controlling the characteristics of the generated game contents by applying latent variable evolution (LVE) \cite{volz2018evolving,giacomello2019searching}.
In these approaches, the game content generator is modeled as a map from a real-valued latent vector to a game content, and is trained on an existing dataset.
The latent vector is optimized such that the generated game contents have the desired features, which can be evaluated through a simulation using AI or human game players.
For example, Volz et al.~\cite{volz2018evolving} first trained a GAN to generate game levels for SMB and then applied LVE to control the difficulty of the generated levels, such as the number of jumps that an AI player can make.
With DGM and LVE, one can generate game contents that satisfy the constraints imposed in a training dataset implicitly by human designers, while realizing some implicit features that may not exist in the training dataset.

Although there are strict constraints on the game level in that it must be playable, unlike the general targets of DGMs (e.g., audio and images), DGMs have been utilized in loosely constrained game (e.g., SMB) that are playable even if the blocks are placed somewhat haphazardly.
Training high-quality DGMs requires a large amount of data, whereas most games can only train low-quality models that produce locally incomplete levels due to the limited data available.
Therefore, it is difficult for a low-quality generative models to generate levels that satisfy the constraint of being playable~\cite{summerville2018procedural}.
A drawback of previous studies is that they only focused on games with loosely constrained levels, even though many game levels have strict constraints.

In this study, we consider PCGML-based level generation for \emph{Angry Birds}, which is a casual action puzzle game.
\emph{Angry Birds} is often targeted in PCG as a challenging game domain for level generation\footnote{\url{https://aibirds.org/level-generation-competition.html}}.
The game applied in the existing study was a game with tile-based levels, whereas the \emph{Angry Birds} are described by an XML file containing various object types with different sizes, measured coordinates, and rotations.
Therefore, the design patterns of levels are massive compared to tile-based levels, and it is difficult to generate stable levels.
The presence of gravity in the level space, combined with the previous difficulty, makes it challenging to generate playable level.
This is because a slight shift in the position of the blocks can cause the level to collapse immediately after the start of the game due to the force of gravity, making the generated level unplayable.
Earlier studies have focused on tile-based game levels with no gravity constraints; therefore, it is difficult to apply existing methods to this game domain, as described in \Cref{sec:ab}.
To the best of our knowledge, there are no examples of level generation methods where DGMs have been applied to \emph{Angry Birds}, or to games with similar playability constraints.

To deal with the aforementioned difficulties, we propose building a level from the bottom up by gradually dropping objects considering the effect of gravity.
We encode a game level as sequence data, where the time step indicates the order in which the objects are dropped.
The types, horizontal positions, and rotations of the dropped objects are also specified in each time step.
Each combination of this specification is considered as a word, and one sentence corresponds to a level.
In sequence encoding, the vertical positions of objects are automatically determined by the order and height of dropping objects, which drastically improve the stability of levels under the gravity.
A long short-term memory (LSTM)-based VAE is used as the generative model.
Furthermore, word embedding techniques are employed to improve the stability of the generated levels.
Then, LVE is applied to design a stable level with the desired features.
Our implementation is available publicly\footnote{https://github.com/yoshinobc/Level-Generation-for-Angry-Birds-with-Sequential-VAE-and-Latent-Variable-Evolution}.

Three experiments are conducted to demonstrate the usefulness of the proposed approach.
First, we compare the stability of the levels generated by the proposed method with those of various models such as tile-based (image-based) generation models and various encoding methods, demonstrating that our encoding method is effective in generating stable levels with high probability.

Second, we compare the diversity of the levels generated by the existing level generator that is not PCGML-based with those generated by the proposed generator trained on the levels generated by the existing one.
When the diversity of the levels generated is extremely low, it is highly likely that stable levels can be generated.
In this experiment, we have shown that the levels generated by the proposed method can generate a variety of levels while maintaining high stability.
Third, to demonstrate the advantages of using our generative model-based approach, we apply LVE to the trained model.
We successfully optimized the number of the specific types of objects in the level and the objective function computed through the AI agent's play.

\section{Angry Birds}\label{sec:ab}
Our game domain is a clone of \emph{Angry Birds}, called \emph{Science Birds}~ \cite{ferreira_2014_a} for the visualization and simulation of levels.
It is a very active game domain in the field of procedural level generation, where level generation competitions are held every year \cite{stephenson2019the}.

In this game, the player destroys all the pigs present in a level by shooting a limited number of birds with a slingshot.
The game level contains different types of objects with different shapes and sizes.
The objects in this game can be roughly classified into platform , TNT blocks, pig blocks, and regular blocks.

The game level of \emph{Angry Birds} is specified by a list of objects. Each object contains the type (e.g., small square stone), vertical and horizontal coordinates, and angle.
There is gravity in the level, and most types of blocks in the level are affected by gravity.
Hence, blocks must be stacked stably under the gravity, by considering the differences of blocks in terms of size and shape.

\begin{figure}[t]
  \centering
  \includegraphics[width=0.5\hsize]{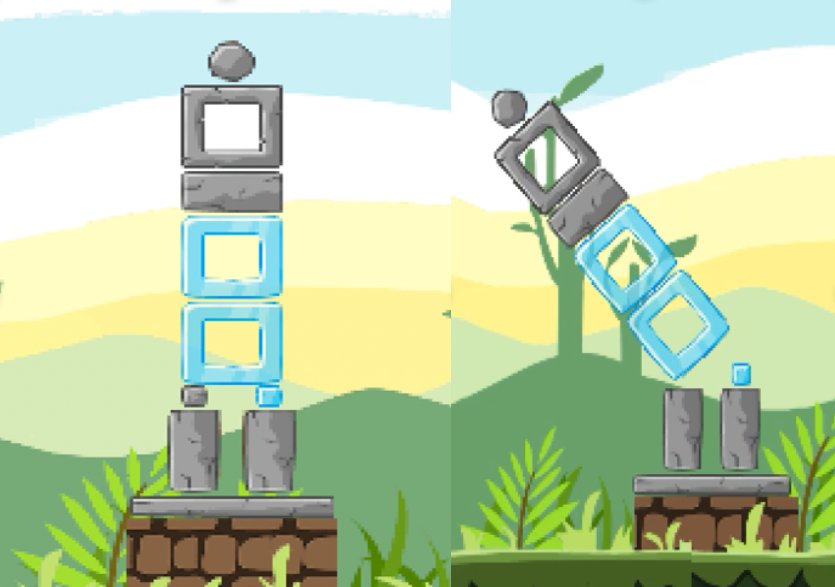}
  \caption{Examples of \emph{Angry Birds} levels. Left: A stable level. Right: Unstable level, where a small stone object is removed from the left figure. This corrupts after the game starts.}
  \label{fig:aibirdslevel}
\end{figure}

There are two main difficulties in developing a ML-based level generation method for \emph{Angry Birds}.
(i) It has a very high degree of freedom in the level design.
The existing level generation techniques that employ DGMs have been applied to tile-based game domains wherein the levels are clearly divided into meshes and all blocks have the same size or the same unit size, and the blocks can be set by specifying the index of the level array.
Therefore, the existing approaches such as \cite{volz2018evolving,torrado2019bootstrapping,thakkar2019autoencoder} encode the tile-based level as an image and successfully use DGMs for image processing.
However, the levels of \emph{Angry Birds} are specified by real numbers, and a large number of meshes are needed to represent the exact stage.
The different block sizes can easily result in overlapping blocks, which may lead to an unplayable level.
(ii) Another difficulty is that the conditions for reaching a playable level are severe because of the existence of gravity. A small error can cause a collapse of the entire level immediately after the game starts, as shown in \Cref{fig:aibirdslevel}.

Existing level generation approaches for \emph{Angry Birds} are designed using domain knowledge, and the search-based approaches are employed.
Ferreira et al.~\cite{ferreira2014a} used a genetic algorithm to find combinations of basic blocks and pre-defined baseline structures that constitute stable levels.
Stephenson et al.~\cite{stephenson2017generating} predefined an algorithm to generate building blocks that are likely to be stable. Changes in pig placement, location of additional blocks, and block materials are updated using heuristics to increase level robustness, and changes in block types to increase level diversity.
This was extended in \cite{stephenson2019agent}, which optimized the parameters of the level generator to maximize the utility computed by the simulation using AI agents.
Abdullah et al.~\cite{abdullah2020generating} proposed an algorithm to create levels that can be destroyed in one shot by combining human-designed baseline structures.
These methods rely on domain knowledge to build stable baseline structures with different objects under the gravity.

\section{Proposed Approach}\label{sec:proposed}
We aimed at developing a DGM-based level generation for game domains where the difficulty in generating stable levels described in \Cref{sec:ab} exists.
We focus on \emph{Angry Birds} as an example of such game domains in this study.

We propose a DGM-based level generation for \emph{Angry Birds}.
The objective is to replace the engineering process of the existing level generation approaches by training a generative model from a dataset and enabling LVE on the top of the trained generator. To utilize an efficient algorithm designed for a general-purpose black-box continuous optimization in the LVE stage, for example, CMA-ES \cite{hansen1996adapting,hansenEC2001,hansenEC2003}, we employ a generative model that uses a continuous latent vector as an input.

\begin{figure}[t]
  \centering
  \includegraphics[width=\hsize]{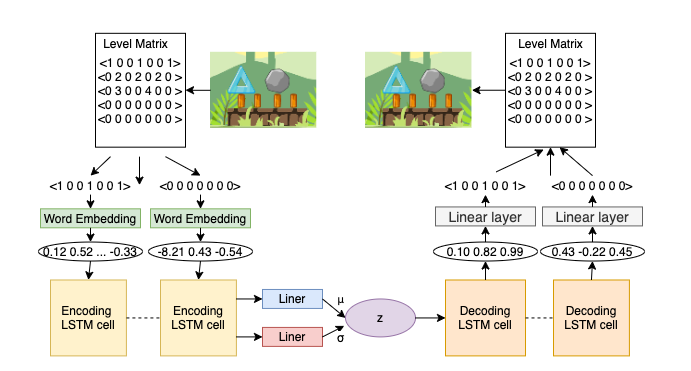}
  \caption{Flowchart of the proposed approach}
  \label{fig:modelarch}
\end{figure}

\Cref{fig:modelarch} shows the overall diagram of the proposed VAE.
Each level is encoded as sequence data (\Cref{sec:encode}).
A sequence is treated as a sentence, and each entry of the sentence is treated as a word (\Cref{sec:word}).
Word embedding is applied (\Cref{sec:embed}). Our generative model is a sequential VAE \cite{bowman2016generating}, which is trained to resemble existing levels in a dataset (\Cref{sec:lstm}).
The LVE explores the latent space of the trained VAE by optimization, for a level with the desired features (\Cref{sec:decode}).
The technical details of the level encoding and decoding are presented in \Cref{apdx:detail-encode,apdx:decode}.

\subsection{Sequential Encoding}\label{sec:encode}

\begin{figure}[t]
  \centering
  \begin{subfigure}{0.5\hsize}%
    \centering%
    \includegraphics[scale=0.25]{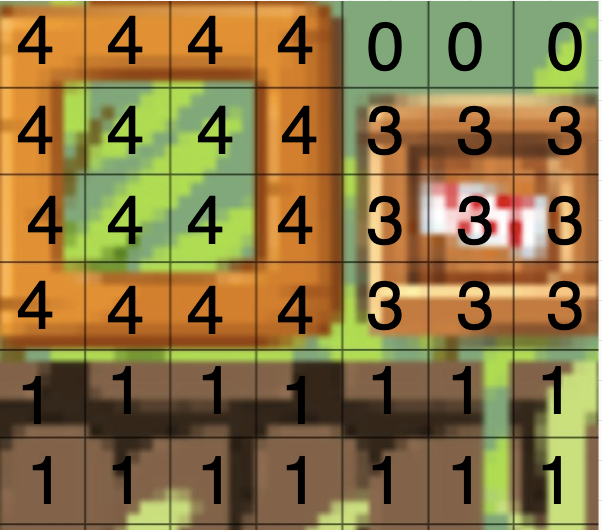}%
    \caption{Dense encoding}%
    \label{fig:juurai}%
  \end{subfigure}%
  \begin{subfigure}{0.5\hsize}%
    \centering%
    \includegraphics[scale=0.25]{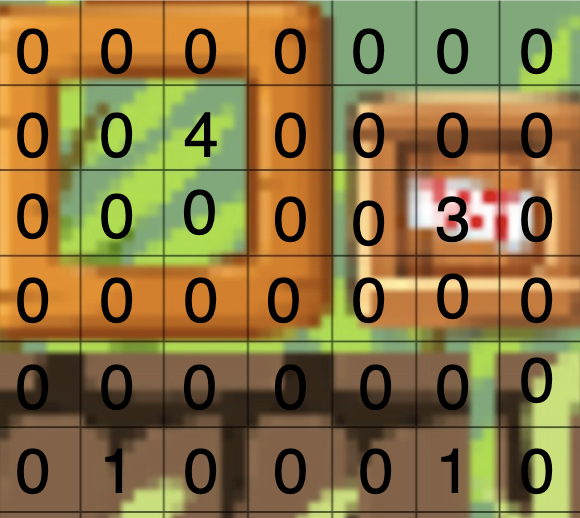}%
    \caption{Sparse encoding}%
    \label{fig:juurai-one}%
  \end{subfigure}%
  \caption{Image-based encoding approaches}
  \label{fig:image-based}
\end{figure}

\providecommand{\ntype}{\texttt{N\_TYPES}}
We propose encoding a level as a sequence data, $\bm{S} = (\bm{s}_1, \dots, \bm{s}_{\maxy})$.
We refer to this as a \emph{level matrix}.
Unlike the tile-based encoding displayed in \Cref{fig:image-based} where a level is divided into a predefined mesh and objects are placed on cells,
our encoding strategy models the level from the bottom to the top.
The time step, $t$, indicates the order of dropping objects, and the type and horizontal positions of the dropping objects are specified by $\bm{s}_t$.
Thus, we incorporate the role of gravity naturally. All objects are forcefully placed on the ground or on other objects.
The maximal time step is set to the maximal length in the training data, which is $\maxy = 30$ in our experiments.

Let $\mathcal{O}$ denote the set of possible object types.
The type of an object is determined by the combination of a type of object (e.g., wood) and its predefined rotation angle. We have $|\mathcal{O}| = \ntype = 61$ types in total.
The horizontal coordinate of a level is divided into a uniform mesh of size $\maxx = 94$ in advance.
In other words, $\bm{s}_t = (o_{t,1}, \dots, o_{t,\maxx})$ for each $t$, and $o_{t, i} \in \mathcal{O}$ is an object dropped at the $i$-th horizontal position at time step $t$.
Because objects in $\mathcal{O}$ are different in size, the vertical coordinates of $o_{t,i}$ and $o_{t,j}$ are not necessarily the same.
These coordinates are determined by physically dropping the specified object at the level containing the objects placed in the previous time steps.

\begin{figure}[t]
  \centering
  \includegraphics[width=0.3\hsize]{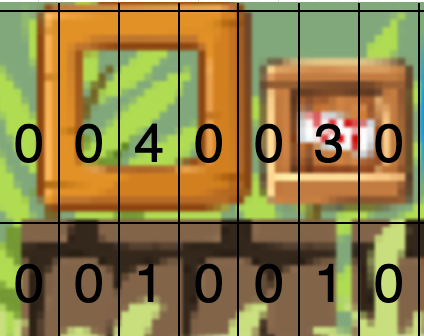}
  \caption{Proposed encoding method}
  \label{fig:konkai}
\end{figure}

\Cref{fig:konkai} depicts this encoding. 
We let $\mathcal{D}_{\bm{S}}$ denote the training dataset of the levels that are processed by the proposed encoding.

\subsection{Word-based Representation}\label{sec:word}

We propose to further process a level matrix, $\bm{S}$, as if it is a natural language sentence.
A level matrix is regarded as a sentence, each row, $\bm{s}_t$, of a level matrix is regarded as a word, and each column, $o_{t,i}$ in $\bm{s}_t$, is regarded as a letter.
The number of alphabets is $\ntype$; a word is composed of $\maxx$ letters.
A level matrix, $\bm{S}$, is mapped to a sequence $\bm{W} = (\bm{w}^{(t)})_{t=1}^{T}$ of one-hot vectors $\bm{w}^{(t)}$ representing a word in the vocabulary.
Vocabulary $\mathcal{W} = \{\bm{s} : \bm{s} \in \bm{S} \text{ for } \bm{S} \in \mathcal{D}_{\bm{S}}\}$ is a set of all rows appearing in the level matrices in the training dataset.
Indexing is performed, and the one-hot vector of a word represents the word index.
The number of words in a sentence is denoted by $T$, and $T = \maxy$ in our case.
Let $\mathcal{D}$ denote the dataset of the word-based sequence representation of the level matrices in $\mathcal{D}_{\bm{S}}$.

The rationale behind the choice of word-based representation is as follows.
In text generation using deep neural networks, \emph{word-based} modeling is more likely to lead to better performance than a \emph{character-based} modeling \cite{kim2016character}.
In character-based modeling, a neural network must learn combinations of characters that exist in the vocabulary, and correlations between words in a sentence.
In our case, a word that does not exist in the vocabulary corresponds to a pattern of blocks that does not appear in a training dataset.
On the one hand, it may be helpful to have a large variation in the generated levels after training.
On the other hand, patterns that do not appear in a training dataset are likely to be invalid (objects that are located too closely can overlap) or unstable.
Based on the above considerations, we employ a word-based approach.

\subsection{Word Embedding}\label{sec:embed}

\providecommand{\LL}{\mathcal{E}}
\providecommand{\UU}{\mathcal{F}}
\providecommand{\DIMX}{\texttt{DIM\_X}}

Word embedding is applied to obtain a real vector representation, $\bm{x}$, of a word $\bm{w}$.
A difficulty in training a VAE for sequence data is its high dimensionality.
A naive representation of a word is a one-hot representation, where the length of the vector is $\lvert\mathcal{W}\rvert$.
It is sparse and high dimensional, and it is difficult to train the VAE adequately for such data.
To easily train the VAE, we utilize word embedding that provides dense and low-dimensional representation.

We employ \texttt{word2vec} with the continuous Bag-of-Words model~\cite{mikolov2013efficient}.
The model is composed of two linear layers, $\LL$ and $\UU$, where $\LL: \bm{w} \mapsto \bm{x}$ maps a word to its corresponding embedding, $\bm{x} \in \mathbb{R}^{\DIMX}$ and $\UU: \mathbb{R}^{\DIMX} \to \mathbb{R}^{\lvert \mathcal{W} \rvert}$.
The conditional probability of $\bm{w}^{(t)}$ given $(\bm{w}^{(t-m)}, \dots, \bm{w}^{(t-1)}, \bm{w}^{(t+1)}, \dots, \bm{w}^{(t+m)})$ is modeled by using $\UU$, where the probability vector is given by the \textsc{softmax} of $\UU\big( \sum_{i=1}^{m} \LL(\bm{w}^{(t-i)}) + \LL(\bm{w}^{(t+i)}) \big)$.
The parameters of $\LL$ and $\UU$ are trained to maximize the conditional probability on a training dataset.
Then, $\bm{x}^{(t)} = \LL(\bm{w}^{(t)})$ is used as the embedding of the word $\bm{w}^{(t)}$.
The sequence of the embedding vectors is denoted by $\bm{X} = (\bm{x}^{(t)})_{t=1}^{T}$.
For simplicity, we considere $\bm{X} = \LL(\bm{W})$ in the following.

\subsection{Sequential Variational Autoencoder}\label{sec:lstm}

\providecommand{\enc}{q_\phi}
\providecommand{\dec}{p_\theta}
\providecommand{\DIMZ}{\texttt{DIM\_Z}}
We model the generation of a sequence $\bm{W}$ that represents a game level as
a variant of the VAE, i.e., sequential VAE \cite{bowman2016generating}.
A VAE consists of two models: an encoder and a decoder.
The encoder is trained to map a high-dimensional feature vector to a distribution of a low-dimensional latent vector, and the decoder is trained to recover an original feature vector from a latent vector.
Both the encoder and decoder networks are modeled by LSTM networks, which can handle sequential data, because our data are sequential data.
The encoder network is a map from a sequence $\bm{X} = (\bm{x}^{(t)})_{t=1}^{T}$ to the mean vector and the log of the coordinate-wise variance of a Gaussian distribution, which models the conditional probability, $\enc(\bm{z} \mid \bm{X})$, of a latent vector, $\bm{z}\in \mathbb{R}^{\DIMZ}$, associated with the given sequence, $\bm{X}$.
The decoder network models the conditional probability, $\dec( \bm{W} \mid \bm{z})$ of a sequence $\bm{W} = (\bm{w}^{(t)})_{t=1}^{T}$ for one-hot word representations, given the latent representation, $\bm{z}$.

The sequential VAE is trained by minimizing the loss function defined by $\beta$-VAE \cite{higgins2017beta}.
The loss function consists of two components: reconstruction loss and Kullback–Leibler (KL) loss.
The reconstruction loss is defined as
\begin{equation*}
  \mathcal{L}_\text{rec}(\mathcal{D}) = \frac{1}{\lvert \mathcal{D} \rvert} \sum_{\bm{W} \in \mathcal{D}} \ln \dec( \bm{W} \mid \bm{z}_{\bm{W}} ) \enspace,
  \quad
  \bm{z}_{\bm{W}} \sim \enc(\cdot \mid \LL(\bm{W}) )
  \enspace.
\end{equation*}
This measures the likelihood of the input sequence to the encoder being reconstructed by the decoder.
The KL loss is defined as
\begin{equation*}
  \mathcal{L}_\text{KL}(\mathcal{D})
  = \frac{1}{\lvert \mathcal{D} \rvert} \sum_{\bm{W} \in \mathcal{D}} D_\text{KL}( \enc(\bm{z} \mid \LL(\bm{W})) \parallel \mathcal{N}(\bm{0}, \bm{I}) ) \enspace,
\end{equation*}
where $D_{KL}$ denotes the KL divergence. This measures the distance from the distribution of latent vectors to the Gaussian prior.
Finally, the loss is defined as the weighted sum, $- \mathcal{L}_\text{rec}(\mathcal{D}) + \beta \cdot \mathcal{L}_\text{KL}(\mathcal{D})$.
Based on \cite{bowman2016generating}, we adopt two techniques---KL annealing and word drop---when training the model. These techniques have been introduced to mitigate the issue in which the trained decoder ignores the latent vector. For details, we refer to \cite{bowman2016generating,higgins2017beta}.

\subsection{Level Generation}\label{sec:decode}

We define our level generator, $\mathcal{G} : \bm{z} \mapsto \bm{W}$, as a deterministic map.
Given a latent vector, $\bm{z}$, $\mathcal{G}(\bm{z})$ is defined by the maximizer, $\bm{W}$, of the trained decoder, $\dec(\bm{W}\mid \bm{z})$.
The levels generated by $\mathcal{G}$ are expected to satisfy \emph{implicit constraints imposed on the training dataset}.
Here, the generated levels are expected to be stable and playable because the training data are stable and playable.

To explore levels that have a desired features, we perform LVE.
Let $f(\bm{W})$ be an objective function. The objective function value of a level $\bm{W}$ can be computed directly from $\bm{W}$, using the results of the AI agents' play or evaluated by human players on a game renderer.
We pose the following optimization problem:
\begin{equation}
  \min_{\bm{z}} f(\mathcal{G}(\bm{z})) \quad \text{subject to} \ \bm{z} \in \mathbb{Z} \enspace,
  \label{eq:lve}
\end{equation}
where $\mathbb{Z} \subseteq \mathbb{R}^{\DIMZ}$ denotes the search space.
Because the generator is trained only for $\bm{z} \in \mathcal{N}(\bm{0}, \bm{I})$, $\mathcal{G}(\bm{z})$ may not be trained for $\bm{z}$, which is unlikely for $\mathcal{N}(\bm{0}, \bm{I})$. Considering this point and the ease of handling constraints, we set $\mathbb{Z} = [-3, 3]^{\DIMZ}$.

\section{Experiments}\label{sec:exp}
We compared the probability of producing a playable levels with generators using different encoding methods.
The comparison results indicated that word-based encoding methods are more suitable than char-based and image-based encoding methods.
Further, we investigated the diversity of levels produced by different generators.
Based on the results of this comparison, we demonstrated that our proposed encoding method generates more diverse levels than other encoding methods.

\subsection{Common setting}

\paragraph{Dataset}
We used a dataset consisting of 200 levels, where 180 levels were considered as the training dataset and 20 levels as the validation dataset.
The validation set was used to adjust the training hyper-parameters.
The dataset size was significantly smaller than those typically used for DGM training; however, it was a realistic number in level generation because datasets used in level generation are created manually.\footnote{In Angry Birds, the size of the level corpus created by Zafar et al.~\cite{zafar2019corpus} was 200. In other games, the size of the dataset used to generate levels for \emph{Lode Runner} was 150~\cite{thakkar2019autoencoder}, and that for SMB was 173~\cite{volz2018evolving}.}
The dataset was generated from IratusAves \cite{stephenson2017generating}\footnote{We used the code provided at \url{https://github.com/stepmat/IratusAves}.}, which was the generator that won the CIG 2018 Level Generation Competition.
The vocabulary size, $\lvert\mathcal{W}\rvert$, in the training dataset was 1503.
Although it may not be very practical, we generated a larger dataset of 10000 levels and trained our model on it for testing; the results are provided in \Cref{apdx:largedataset}.

\paragraph{Model}
The output of the word embedding had a dimension $\DIMX = 50$. The dimension of the latent vector in the VAE was $\DIMZ = 60$. The number of hidden neurons for the encoder and decoder LSTM networks was $400$.
The VAE was trained on the training data with $500$ epochs.
The word drop rate was set to $0.3$, and the KL loss was ignored in $250$ epochs.

\paragraph{Baselines}
We compared the following approaches:
\begin{itemize}
\item Proposed: as described in \Cref{sec:proposed}
\item W/o WE: Word-based encoding without word embedding
\item Char(D): Character-based dense encoding
\item Char(S): Character-based sparse encoding
\item Image(D): Image-based dense encoding
\item Image(S): Image-based sparse encoding
\end{itemize}
In dense encoding (\Cref{fig:juurai}), cells corresponding to an object are filled with the object index, whereas in sparse encoding, the object index is located only at the center cell of the object (\Cref{fig:juurai-one}).
In the image-based encoding, a level is divided into a uniform mesh, and each cell has an object type index.
We employed WGAN-GP \cite{gulrajani2017improved} for the image-based approaches, that were used in the existing level generator \cite{volz2018evolving}. In WGAN-GP, the row and column sizes of a level were set as 96.
There was no difference between dense and sparse encoding in word-based encoding because they were transformed into word indices.
For image-based encoding approaches, an imperfect level matrix (overlapping objects and missing and extra cells for an object) can be repaired to some extent by manually designed procedure.
\begin{table}[t]
  \begin{center}
    \caption{Numbers of unique uni-gram and bi-gram types present in the levels generated by different methods.}
    \small
    \begin{tabular}{cccccc} \toprule
      & Training & \bf{Proposed} & W/o WE & Char(D) & Char(S) \\ \midrule
      Uni-gram & 1503 & 1376  & 233 & 88 & 426\\
      Bi-gram & 1832 & 2737 & 267 & 93 & 848\\ \bottomrule
    \end{tabular}
    \label{tab:gram-num}
  \end{center}
\end{table}
\providecommand{\fsize}{0.27}
\begin{figure*}[!t]
  \centering
  \begin{subfigure}{\fsize\hsize}%
    \centering%
    \includegraphics[width=\hsize]{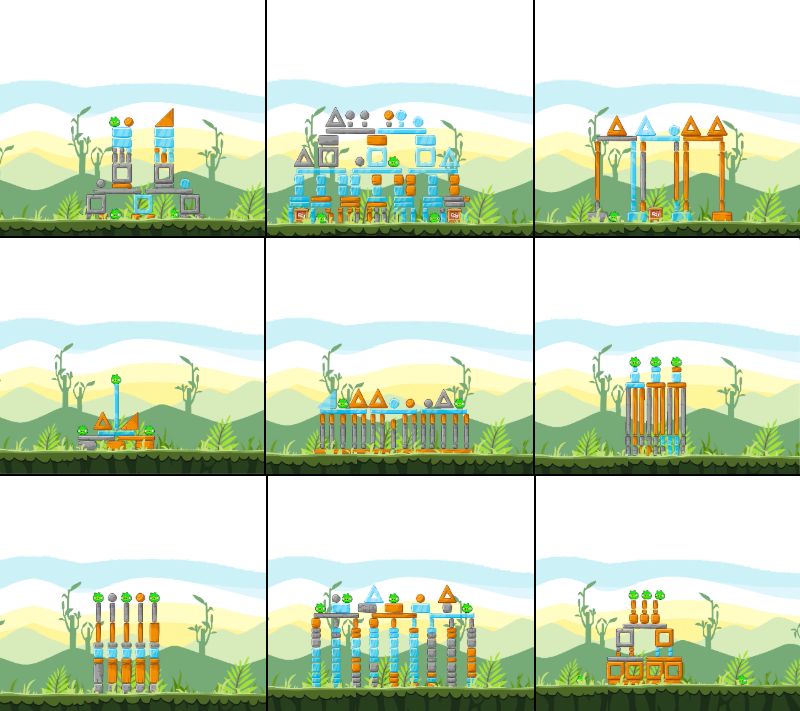}%
    \caption{Proposed}%
    \label{fig:proposed_level}%
  \end{subfigure}%
  \
  \begin{subfigure}{\fsize\hsize}%
    \centering%
    \includegraphics[width=\hsize]{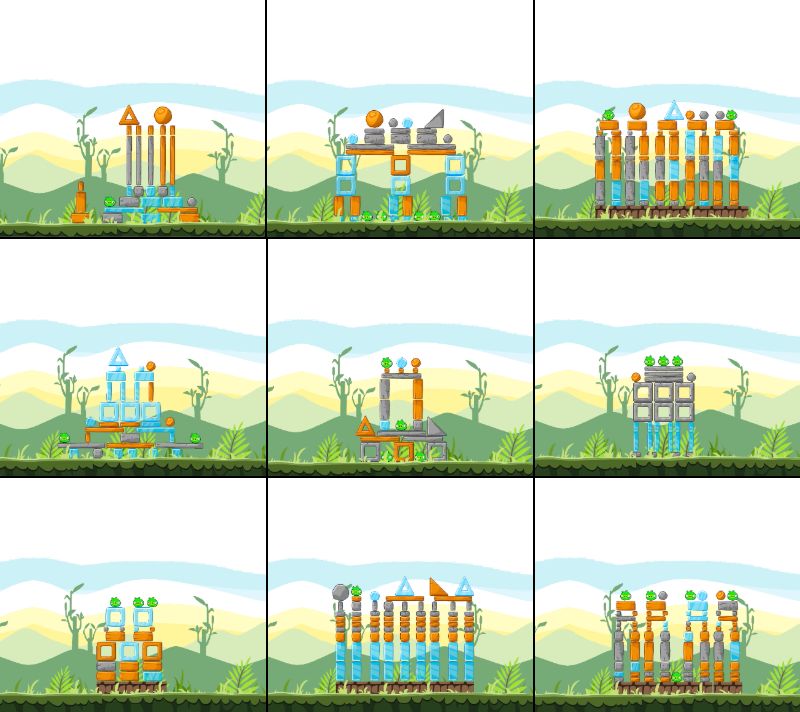}%
    \caption{W/o WE}%
    \label{fig:wowe_level}%
  \end{subfigure}
  \begin{subfigure}{\fsize\hsize}%
    \centering%
    \includegraphics[width=\hsize]{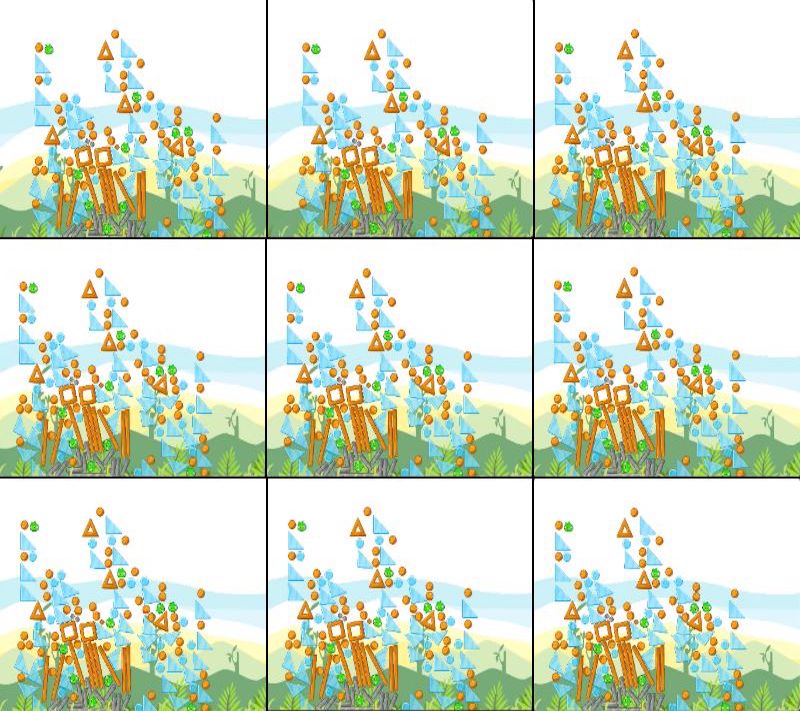}%
    \caption{Char(D)}%
    \label{fig:chard}%
  \end{subfigure}
  \\
  \begin{subfigure}{\fsize\hsize}%
    \centering%
    \includegraphics[width=\hsize]{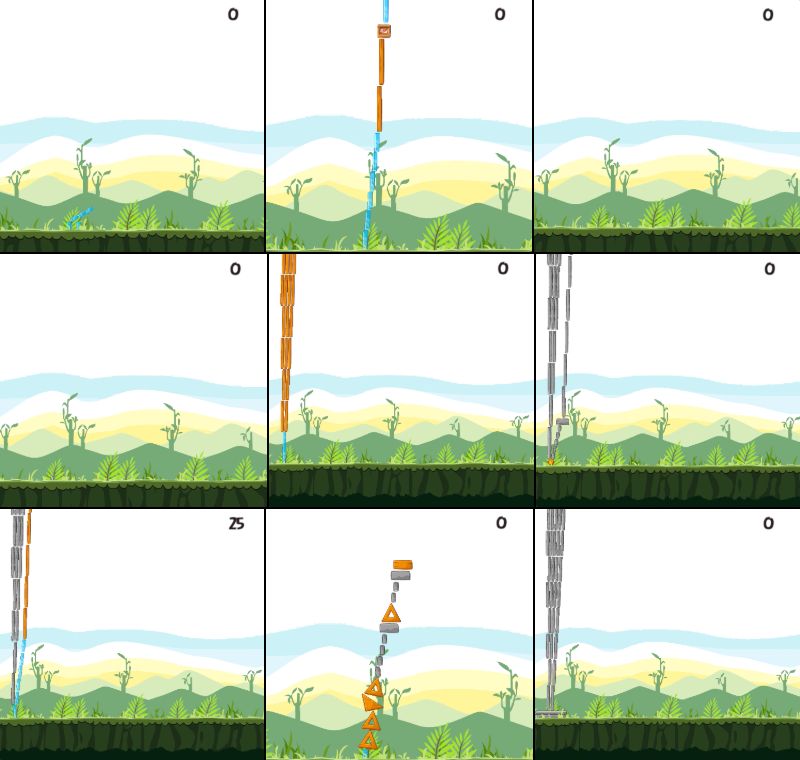}%
    \caption{Char(S)}%
    \label{fig:chars}%
  \end{subfigure}%
  \
  \begin{subfigure}{\fsize\hsize}%
    \centering%
    \includegraphics[width=\hsize]{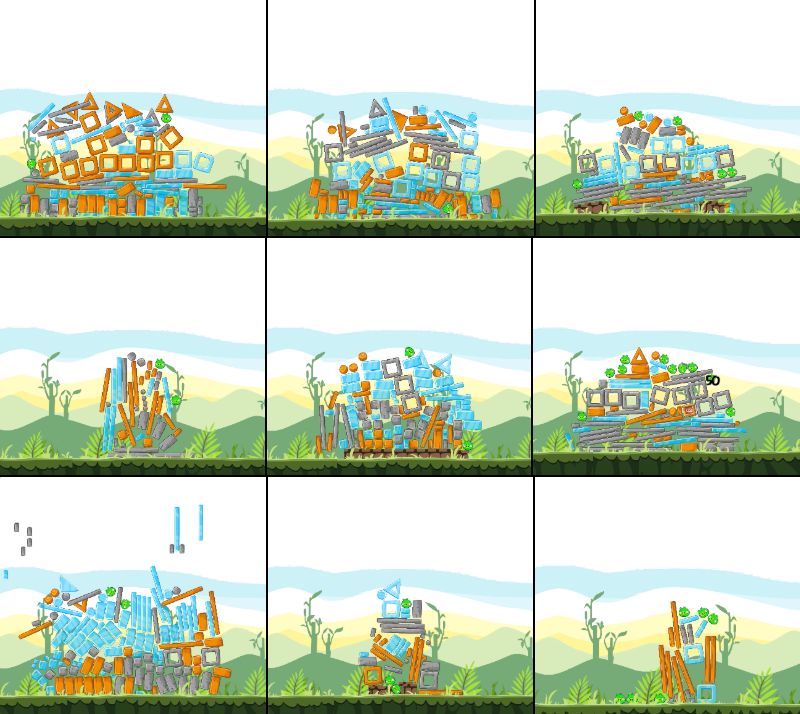}%
    \caption{Image(D)}%
    \label{fig:image_d}%
  \end{subfigure}
  \begin{subfigure}{\fsize\hsize}%
    \centering%
    \includegraphics[width=\hsize]{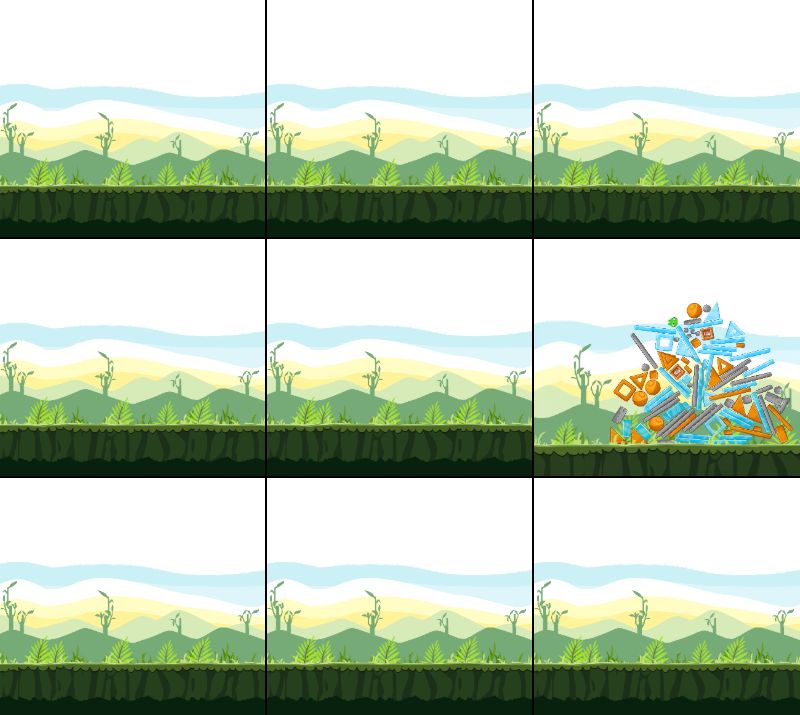}%
    \caption{Image(S)}%
    \label{fig:image_s}%
  \end{subfigure}
  \caption{Levels generated by different generators}
  \label{fig:enc-all-level}
\end{figure*}

\begin{table}[t]
  \begin{center}
    \caption{Number of stable levels among 100 levels generated from each generator.
      We consider a level to be stably generated if it does not fall after the game starts.}
    \small
    \begin{tabular}{cccccc} \toprule
       \bf{Proposed} & W/o WE & Char(D) & Char(S) & Image(D) & Image(S)      \\ \midrule
          \bf{96}   &  93 &  0 &  10 & 6 & 0           \\ \bottomrule
    \end{tabular}
    \label{tab:stable-rate}
  \end{center}
\end{table}

\subsection{Diversity}\label{sec:diversity}
We compared the diversity of the level matrices computed from the training dataset and those generated by the proposed method, W/o WE, Char(D), and Char(S).

A level matrix was generated by sampling a latent vector from the multivariate normal distribution, $\mathcal{N}(\bm{0}, \bm{I})$, and feeding it to the trained decoder. We produced a 1000 level matrix using each method.

We measured the diversity of level matrices using n-gram, which is used in natural language processing.
An n-gram is a contiguous sequence of $N$ words or characters.
Since our generator is word-based, we use word as the unit.
The diversity measure for n-grams is called distinct-N (a diversity-promoting objective function for neural conversation models).
We counted the number of unique unigrams ($N = 1$) and bigrams ($N = 2$).
The higher the number of unigrams present in the generated levels, the richer are the word (a group of blocks dropped at once in a single trial) patterns are in the generated levels.
The more bigrams there are in the generated levels, the higher is the number of levels that can be generated with various combinations of words.

\Cref{tab:gram-num} summarizes the numbers of unique unigrams and unique bi-grams included in the training dataset and $1000$ level matrices generated by the trained generators.
Word-based encoding cannot generate unigrams that do not appear in the training dataset; however, it is possible to generate bi-grams that do not appear in the training dataset.

The results indicated that approximately $92\%$ of uni-gram types can be reproduced, and that the variation of bigrams was as high as $149\%$ of the training dataset.
These values, especially those for bigrams, may even increase as the number of generated levels increases.
For W/o WE, char(S), char(D), the diversity of the generated levels was very low for both unigrams and bigrams, probably because of the difficulty in learning.
\subsection{Stability}\label{sec:level-enc-ex}

We evaluated the stability of the levels generated by the proposed level generator and those generated using different encoding approaches.
We verified our expectations: 1) the proposed sequential encoding approach is advantageous in generating stable levels over image-based encoding approaches and 2) the word-based encoding approaches produce more stable levels than character-based encoding approaches.
Stability is measured by the probability of successfully generating a level that does not collapse without any action.

\Cref{tab:stable-rate} lists the number of stable levels among the 100 levels generated by each approach.
Sequential encoding drastically improved the stability of the generated levels over image-based approaches.
The proposed approach succeeded in generating stable levels with a probability of about $96\%$.
Further, W/o WE generated stable levels with a high probability of $93\%$.
The levels produced by the generators of character-based and image-based encoding methods were not stable.

\Cref{fig:enc-all-level} shows example levels generated by these approaches.
The levels generated by image-based approaches were unstable.
Character-based approaches often failed to stack objects (objects moved left or right as they were stacked on each other).
Here, we observed the difficulty in training the correlation between distant characters.
The levels were more likely to be stable for word-based approaches, especially for the proposed approach.

\subsection{Discussion}
The results of the aforementioned experiments indicated that our proposed generator and W/o WE---both word-based encoding strategies---generated more stable levels than char-based (Char(D) and Char(S)) and image-based (Image(S) and Image(D)).
Further, the diversity of the levels generated by the proposed approach was higher than that generated by other sequence-based encoding strategies (W/o WE, Char(D), and Char(S)).

In the experiment described in \Cref{sec:diversity}, character-based embedding can theoretically generate words that do not appear in the training dataset; hence, the number of unique unigrams of Char(D) may be larger than that of the training dataset. However, in this experiment, the number of unigrams and bigrams was smaller than that in the training dataset. This is probably because the dataset size was relatively small compared to those typically used to train DGMs, which resulted in an unsuccessful learning procedure. \Cref{apdx:largedataset} details the results on the larger dataset.

Word embedding had the effect of dimensional compression. Because the dataset used in this study was small (the number of dimensions is 1503), it was not too large even without word embedding. Therefore, in this experiment, there was no considerable difference between the model without word embedding and the generator with word embedding. However, on the large dataset outlined in \Cref{apdx:largedataset}, the level of stability was lower without word embedding.
For large datasets, the loss during the training of the sequential VAE for the proposed method was approximately $10^3$ times smaller than that for the W/o WE.
This implied that the word embedding makes model training more tractable.

In the experiment described in \Cref{sec:level-enc-ex}, generators of character-based and image-based encodings unlike those of word-based encodings could rarely produce stable levels.
Word-based encoding was different from char-based encoding in that the combinations of blocks were determined in advance as words.
Because word-based encoding determines only the order of combination, it can generate stable levels.
However, char-based encoding determines the combination of blocks without using the word.
Therefore, compared to word-based encoding, char-based encoding generated fewer stable levels.
Unlike the image-based encoding method, the word-based encoding method generated levels by dropping blocks from the top.
This solved the problems of overlapping blocks and blocks floating apart, thereby achieving stability for word-based encoding compared to that for image-based encoding.

\section{Latent Variable Evolution} \label{sec:lve}
An advantage of using DGMs for level generation is that LVE~\cite{volz2018evolving,giacomello2019searching} can be employed after training the generator to identity levels with certain characteristics.
However, existing studies on sequential VAE have reported that the latent vector tends to be ignored, and the variation of outputs is created by the randomness of $\mathcal{G}$ itself \cite{roberts2018hierarchical}.
Thus, $\mathcal{G}$ is not suitable for LVE, because we cannot control the features of the generated levels.
Although we adopted techniques to mitigate this issue in training the sequential VAE, the applicability of LVE with the proposed level generator must be evaluated.
In this section, we demonstrate the applicability by performing LVE with four different objectives.

\begin{figure}[t]
  \begin{subfigure}{0.5\hsize}%
    \centering%
    \includegraphics[width=\hsize]{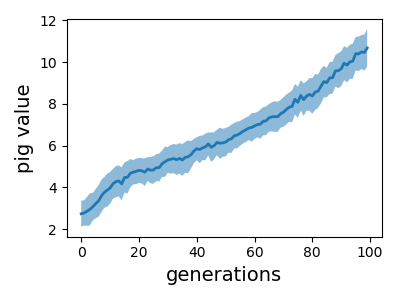}%
  \end{subfigure}%
  \begin{subfigure}{0.5\hsize}%
    \centering%
    \includegraphics[width=\hsize]{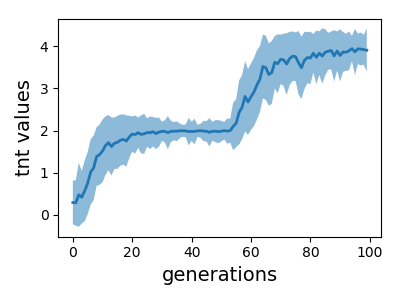}%
  \end{subfigure}
  \caption{Single run results of LVE without simulation. Average (solid line) and standard deviation (band) of $N_\text{Pigs}$ (left) and $N_\text{TNTs}$ (right) over ten levels generated using the best solution for each generation.}
  \label{fig:lvepigtnt}
\end{figure}

\subsection{LVE without Simulation}\label{sec:lvewo}
The first experiment was to maximize the number of specific blocks, namely, i.e., the number $N_\text{Pigs}$ of pig blocks and the number $N_\text{TNTs}$ of TNT blocks.
These are the most important blocks in \emph{Angry Birds} and determine the the game's interest.

In this experiment, instead of directly optimizing the latent vector, $z$, by solving \eqref{eq:lve}, we parameterized the latent vector with $\alpha \in \mathbb{R}$ and $\bm{\beta} \in \mathbb{R}^{\DIMZ}$ as
\begin{equation*}
  \bm{z} \sim \mathcal{N}(\bm{\beta}, \alpha \cdot \bm{I}) \enspace.
\end{equation*}
After optimization, we utilized the solution, $(\alpha^*, \bm{\beta}^*)$, to generate multiple latent vectors, $\bm{z}_1,\dots, \bm{z}_m$, which were considered as input for the generator, thus leading to a variety of levels, $\mathcal{G}(\bm{z}_1), \dots, \mathcal{G}(\bm{z}_m)$. Therefore, this formulation allowed variation in levels after the LVE process.
We minimized the following objective function:
\begin{equation*}
  F(\alpha, \bm{\beta}) = \frac1m \sum_{i=1}^{m} f (\mathcal{G}(\bm{z}_i)) \quad \bm{z}_i \sim \mathcal{N}(\bm{\beta}, \alpha \cdot \bm{I}) \enspace,
\end{equation*}
where $f$ denotes the negative of $N_\text{Pigs}$ or the negative of $N_\text{TNTs}$.
We set $m = 30$ in this experiment.

\begin{figure}[t]
  \begin{center} 
    \centerline{\includegraphics[width=\hsize]{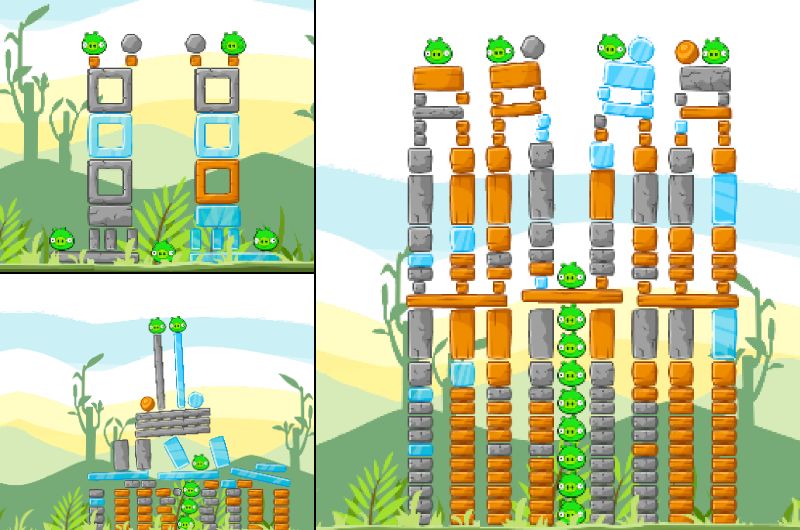}}
    \caption{Results of the maximization of $N_\text{Pigs}$. Levels generated at the 20th (top left), 40th (bottom left), and 100th (right) generations are displayed.
      There are 5, 6, and 13 pig blocks in the levels, respectively.}
    \label{fig:lvepiglevel}
  \end{center}
\end{figure}

CMA-ES was used to optimize $\alpha$ and $\bm{\beta}$ with a population size of $60$.
We set the search space for $(\alpha, \bm{\beta})$ to $[0, 2] \times [-3, 3]^{\DIMZ}$.
The software library pycma~\cite{hansen2019pycma} was used.
We used the proposed generator, $\mathcal{G}$, trained in \Cref{sec:exp}.

\Cref{fig:lvepigtnt} shows the results of the optimization.
The average and standard deviation of $N_\text{Pigs}$ and $N_\text{TNTs}$ were computed at $10$ levels, $\mathcal{G}(\bm{z}_1), \dots, \mathcal{G}(\bm{z}_{10})$, using independent and $\mathcal{N}(\bm{\beta}_t^*, \alpha_t^* \cdot \bm{I})$ distributed random vectors, $\bm{z}_1, \dots, \bm{z}_{10}$, where $(\alpha_t^*, \bm{\beta}_t^*)$ was the best solution generated at generation $t$.
We observed that the numbers of these blocks were increased successfully.
This implied that the trained generator does not ignore the latent vectors, and the features (here, $N_\text{Pigs}$ and $N_\text{TNTs}$) can be controlled by the LVE.

\Cref{fig:lvepiglevel} shows the levels generated during the optimization.
Here, two points need to be considered.
First, although our objective was to maximize the number of pig blocks, the number of other blocks increased as well. This is because the training dataset did not include levels that contained pig blocks alone. This is promising because we would like to generate a level that can be naturally compared to the training dataset but has a features that are not necessarily included in the dataset.
Second, the levels generated by the LVE process were more or less stable.
However, there are levels produced during the LVE process that are not fully stable, such as the level generated at the 40th generation in \Cref{fig:lvepiglevel}.
Although the generator was trained to output stable levels, employing a constraint-handling technique to penalize unstable levels during the LVE process may be desired.

\subsection{LVE with Simulation}

Next, we demonstrated the LVE using a simulation with an AI agent\footnote{\url{https://gitlab.com/aibirds/sciencebirdsframework}}.
We designed two objective functions that were computed through the AI agents play.

The first objective function aimed at measuring the difficulty of a level was computed as
\begin{equation}
  \label{eq:sim1}
   \max(5 - N_{\mathrm{Birds}}, N_{\mathrm{RemBirds}}) \times 10 + N_{\mathrm{Blocks}} \enspace,
\end{equation}
where $N_{\mathrm{Birds}}$, $N_{\mathrm{Blocks}}$, and $N_{\mathrm{RemBirds}}$ represent the numbers of birds at the level at the start of the game, blocks at the level at the start of the game, and birds at the level after the game ends, respectively.
The number of birds at the start of the game is automatically determined by the number of pigs and blocks present at the level, which is similar to that in IratusAves.
We expected the optimized level to have $N_\mathrm{Birds} \geq 5$, and they should all be used by the AI agent ($N_\mathrm{RemBirds}$). The number of blocks in the level was minimized; however, the priority was placed on the first term (because of the coefficient of $10$).

The second objective function aimed at measuring the aesthetics of a level by determining how well the level retained its original shape when all pig blocks were destroyed; the level was then cleared, which was calculated by
\begin{equation}
  \label{eq:sim2}
  \max(60 - N_{\mathrm{Blocks}}, N_{\mathrm{Blocks}} - N_{\mathrm{RemBlocks}}) \times 10 - N_{\mathrm{Pigs}} \enspace.
\end{equation}
where $N_{\mathrm{RemBlocks}}$ represents the number of blocks at the level after the game ends.

CMA-ES with a boundary constraint implemented in \texttt{pycma} was used to optimize \eqref{eq:lve} with a population size of $60$.
We used the proposed generator, $\mathcal{G}$, trained in \Cref{sec:exp}.

\begin{figure}[t]
  \begin{subfigure}{0.5\hsize}%
    \centering%
    \includegraphics[width=\hsize]{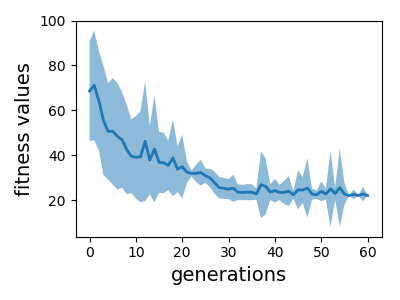}%
  \end{subfigure}%
  \begin{subfigure}{0.5\hsize}%
    \centering%
    \includegraphics[width=\hsize]{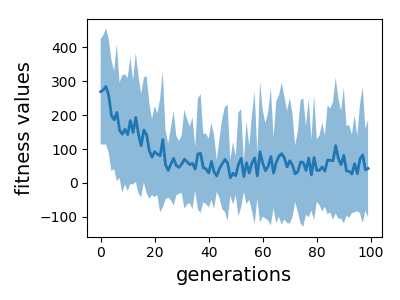}
  \end{subfigure}
  \caption{Single run results of LVE with simulation. The average (solid line) and standard deviation (band) of the objective values, \Cref{eq:sim1} (left) and \Cref{eq:sim2} (right), over a population of $60$ for each generation are displayed.}
  \label{fig:lvesim}
\end{figure}

\begin{figure}[t]
  \begin{subfigure}{0.5\hsize}%
    \centering%
    \includegraphics[width=\hsize]{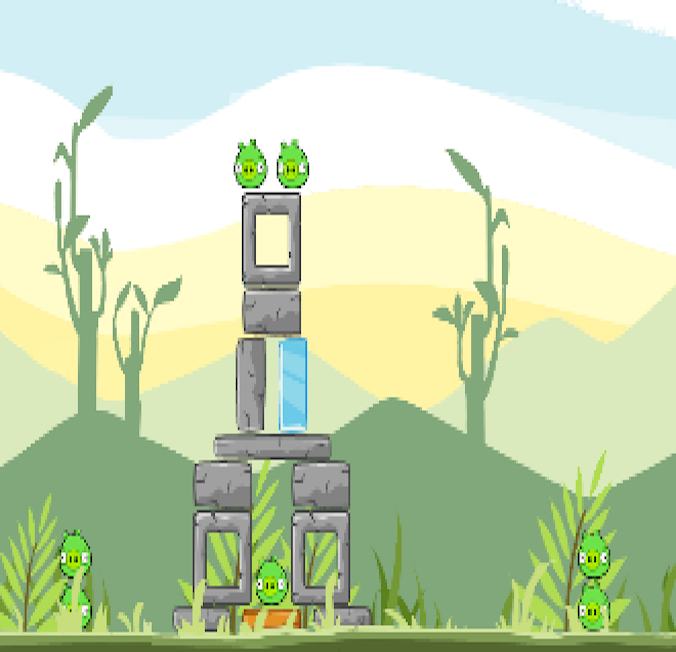}%
    \caption{\Cref{eq:sim1}}%
    \label{fig:sim1_level}%
  \end{subfigure}%
  \begin{subfigure}{0.5\hsize}%
    \centering%
    \includegraphics[width=\hsize]{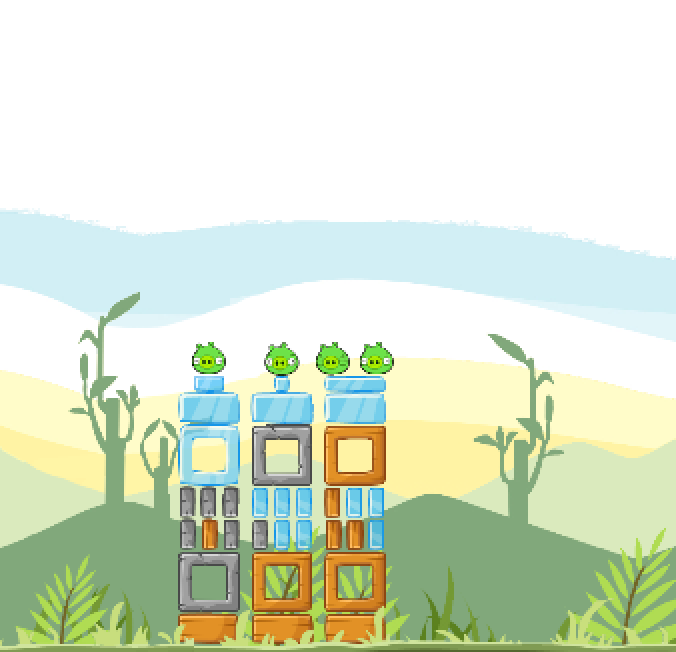}%
    \caption{\Cref{eq:sim2}}%
    \label{fig:sim2_level}%
  \end{subfigure}
  \caption{Results of minimization of \Cref{eq:sim1} (left) and \Cref{eq:sim2} (right). Levels generated at the 60th and 100th generation are displayed.}
  \label{fig:sim1sim2_level}
\end{figure}

\begin{figure}[t]
  \centering%
  \includegraphics[scale=0.6]{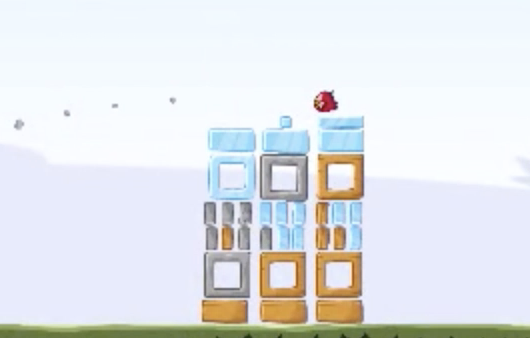}%
  \caption{Level after the AI agent's play on \Cref{fig:sim2_level}.}%
  \label{fig:sim2_level_play}%
\end{figure}

The results are presented in \Cref{fig:lvesim}. The average and standard deviation of the objective values over $60$ candidate solutions generated at each iteration are displayed.
\Cref{fig:sim1sim2_level} shows the levels generated at the last generation (\Cref{eq:sim1} is $60$ generations; \Cref{eq:sim2} is $100$ generations).
The objective functions were successfully minimized, and the generated levels were stable; although subjective, they looked natural.
As shown in \Cref{fig:sim1_level}, the objective was reflected in the generated level.
\Cref{eq:sim1} was designed to minimize the number of birds left after the AI agent plays.
Further, the number of blocks was minimized, which was expected to result in a simple but challenging level.
In fact, at this level, the pigs were grouped in four discrete positions, and the number of birds was also four, so it is probably necessary to use all the birds to successfully shoot all the pigs.
For \Cref{fig:sim2_level}, the level was optimized to minimize the blocks that the birds have broken after the AI agent plays the game.
In fact, the level after the agent has played the game showed that most of the blocks were unbroken, as shown in \Cref{fig:sim2_level_play}.

\section{Conclusion and Future work}
We proposed a level generator for \emph{Angry Birds}.
To overcome the difficulties in generating a level that is stable under gravity, we encoded the level as sequence data and train a sequential VAE for level generation.
The experimental results showed that the trained generator can generate unigrams as diverse as the levels of the training data, levels with more bigrams than the training data, and stable levels with high probability.
We developed a level generator for \emph{Angry Birds}, in which, it is difficult to generate playable levels using existing DGMs.
Unlike existing non-ML-based level generators that target \emph{Angry Birds}, we did not rely on domain knowledge.
We expect that this approach can be generalized to other game domains with gravity at this level.

Two possible directions to extend the proposed approach are as follows.
One is designing a constraint-handling technique to penalize unstable levels during the LVE process.
Although the proposed level generator is trained to output stable levels when the latent vector is generated from the normal distribution, the output can be unstable when performing LVE, as shown in \Cref{fig:lvepiglevel}.
The other direction is to develop a methodology to evaluate the level in a human-in-the-loop manner.
Designing an objective function to reflect a qualitative measure such as aesthetics is a difficult task.

The PCG is not only useful for game content designers, but also for improving the robustness of AI agents based on reinforcement learning.
In reinforcement learning, it has been reported that if the environment used for training is fixed, it will overfit the specific training environment and fail to generalize to the modified environment \cite{justesen2018illuminating}.
A promising approach to learn robust policies is to train agents at many different game levels \cite{cobbe2019quantifying}.
The proposed level generator was used to train a robust \emph{Angry Birds} agent.
Moreover, we expect that LVE will allows us to locate the weakness of a learning agent and actively train it by generating levels at which the agent is poor.
This is another direction for our future work.

\begin{acks}
  This work was partially supported by JSPS KAKENHI Grant Number 19H04179.
\end{acks}

\balance
\bibliographystyle{ACM-Reference-Format}
\bibliography{acs-paper}

\clearpage
\appendix

\section{Details of Encoding}\label{apdx:detail-encode}

The details of our encoding method is summarized in \Cref{alg:encode}.
The input to the encoding algorithm is a list of objects whose elements include the object name (kind), $x$ (horizontal), and $y$ (vertical) coordinates, and rotation angle $\psi$, which are specified in the XML format.
The output is a matrix of size $\maxy \times \maxx$.
Each column index corresponds to a horizontal coordinate where an object drops, and each row index represents the order of an object drop in each column. An integer value in each element of the matrix indicates the object type ($0$ for nothing to drop).

\begin{algorithm}[h]
  \caption{Encode}
  \label{alg:encode}
  \begin{algorithmic}[1]
    \renewcommand{\algorithmicrequire}{\textbf{Input:}}
    \renewcommand{\algorithmicensure}{\textbf{Output:}}
    \REQUIRE ObjectList (Object = [name, $x$, $y$, $\psi$])
    \ENSURE LevelMatrix
    \FOR {row = 0 to \maxy}
    \FOR {col = 0 to \maxx}
    \STATE LevelMatrix[row][col] = name2ind(``Space'')
    \ENDFOR
    \ENDFOR
    \STATE ObjectList = sortAscendingY(ObjectList)
    \STATE $y$ , row = \groundy, 0
    \FOR {Obj in ObjectList}
    \IF {Obj.$y$ - $y \geq \eps$}
    \STATE row += 1
    \ENDIF
    \STATE $y$ = Obj.$y$
    \STATE col = float2ind(Obj.$x$)
    \STATE LevelMatrix[row][col] = name2ind(Obj.name, Obj.$\psi$)
    \ENDFOR
    \RETURN LevelMatrix
  \end{algorithmic}
\end{algorithm}

Lines 1--5 initialize all the elements of the level matrix with $0$.
Line 6 sorts the object list in the ascending order of the $y$-coordinate of the object.
Line 7 initializes the $y$ value to the $y$-coordinate of the ground, denoted by \groundy.
Lines 8--15 set the objects in the input object list to the level matrix.
If the $y$-coordinate of the current object is equal to the current $y$ value up to $+\eps$, where a small number $\eps = 0.1$ is introduced to treat numerical errors in floating point numbers, we place the object in the same row as the previous ones. Otherwise, the row index is incremented.
Line 13 determines the index of the discretized $x$-coordinate to place the object.
Line 14 sets the object type in the corresponding element in the matrix.

\section{Details of Decoding}\label{apdx:decode}

A sequence, $\bm{W} = (\bm{w}^{(t)})_{t=1}^{T}$, which is the output of the decoder, is finally converted into an XML file.
First, we transform $\bm{W}$ into the corresponding level matrix by transforming the word representation, $\bm{w}^{(t)}$, into the corresponding integer vector.
Then, the level matrix is parsed to an XML, by performing the reverse encoding process.

The details of the decoding algorithm are shown in \Cref{alg:decode}.
Line~1 initializes the object list.
The \platformy, which is initialized in Line 2, represents the $y$-coordinate of the highest platform that exists at the current level.
The objects in the level matrix are added to the object list in Lines 2--21.
Line 5 finds the name and rotation angle of the object from the elements of the level matrix.
The $y$-coordinate of the object is calculated in Lines 8--13. Line 8 initializes the $y$-coordinate of an object with the height of the ground.
In Line 10, whether the object already added in the object list is under the object that we want to place now is checked by is\_under.
Furthermore, is\_under always returns True when the ``name'' variable of the argument is ``Platform.''
In Line 11, calc\_y calculates the $y$-coordinate of the object to be placed above other objects already placed in the same $x$-coordinate.
In Line 15, update\_plaftorm\_y updates \platformy\ when the ``name'' variable of the argument is ``Platform.''
Line 17 adds the object to the object list, and Line 18 sorts the object list in the ascending order of the $y$-coordinate.

\begin{algorithm}
  \caption{Decode}
  \label{alg:decode}
  \begin{algorithmic}[1]
    \renewcommand{\algorithmicrequire}{\textbf{Input:}}
    \renewcommand{\algorithmicensure}{\textbf{Output:}}
    \REQUIRE LevelMatrix
    \ENSURE ObjectList
    \STATE ObjectList = []
    \STATE \platformy = \groundy
    \FOR {row = 0 to \maxy}
    \FOR {col = 0 to \maxx}
    \STATE name, $\psi$ = ind2name(LevelMatrix[row][col])
    \IF {name != ``Space''}
    \STATE $x$ = ind2float(col)
    \STATE $y$ = \groundy
    \FOR{object in ObjectList}
      \IF{is\_under(object, name, $x$, $\psi$)}
        \STATE $y$ = calc\_y(object, name, $x$, $\psi$)
      \ENDIF
    \ENDFOR
    \IF {name == ``Platform''}
      \STATE \platformy = update\_platform\_y($y$, \platformy)
    \ENDIF
    \STATE ObjectList.append([name, $x$, $y$, $\psi$])
    \STATE ObjectList = sortAscendingY(ObjectList)
    \ENDIF
    \ENDFOR
    \ENDFOR
    \RETURN ObjectList
  \end{algorithmic}
\end{algorithm}

\section{Large Dataset Experiments}\label{apdx:largedataset}
\begin{figure*}[!t]
  \begin{subfigure}{0.33\hsize}%
    \centering%
    \includegraphics[width=\hsize]{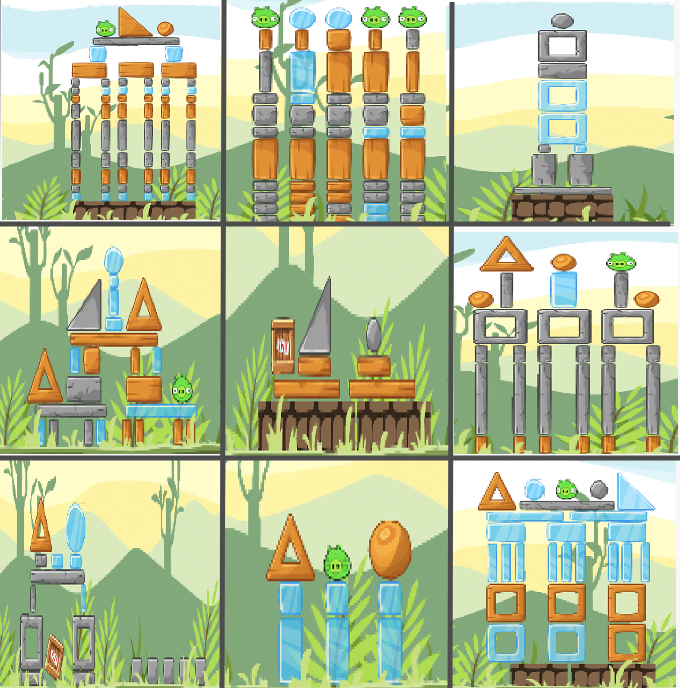}%
    \caption{Proposed}%
    \label{fig:proposed_level_l}%
  \end{subfigure}%
  \begin{subfigure}{0.33\hsize}%
    \centering%
    \includegraphics[width=\hsize]{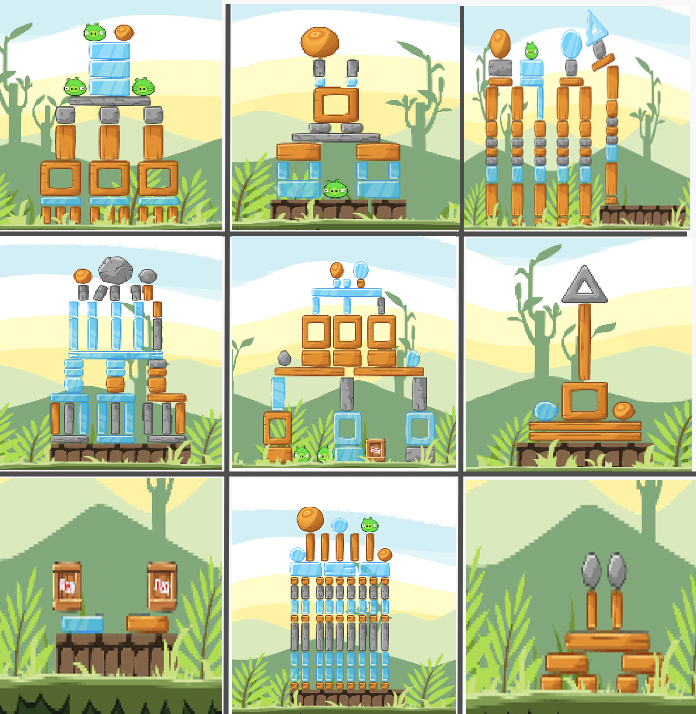}%
    \caption{W/o WE}%
    \label{fig:wowe_level_l}%
  \end{subfigure}
  \begin{subfigure}{0.33\hsize}%
    \centering%
    \includegraphics[width=\hsize]{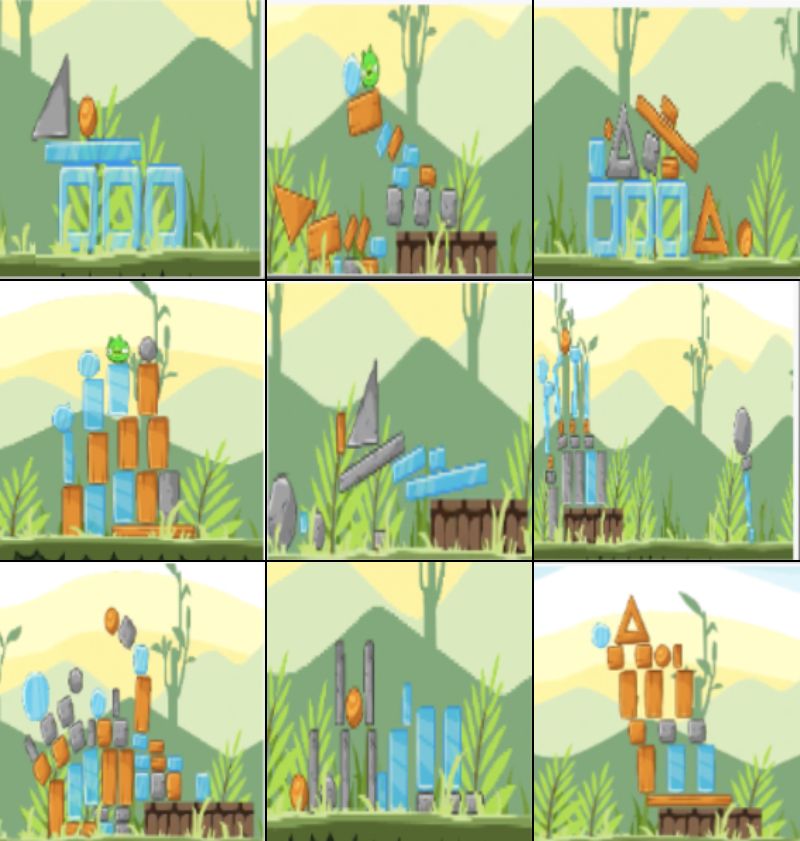}%
    \caption{Char(D)}%
    \label{fig:chard_l}%
  \end{subfigure}
  \\
  \begin{subfigure}{0.33\hsize}%
    \centering%
    \includegraphics[width=\hsize]{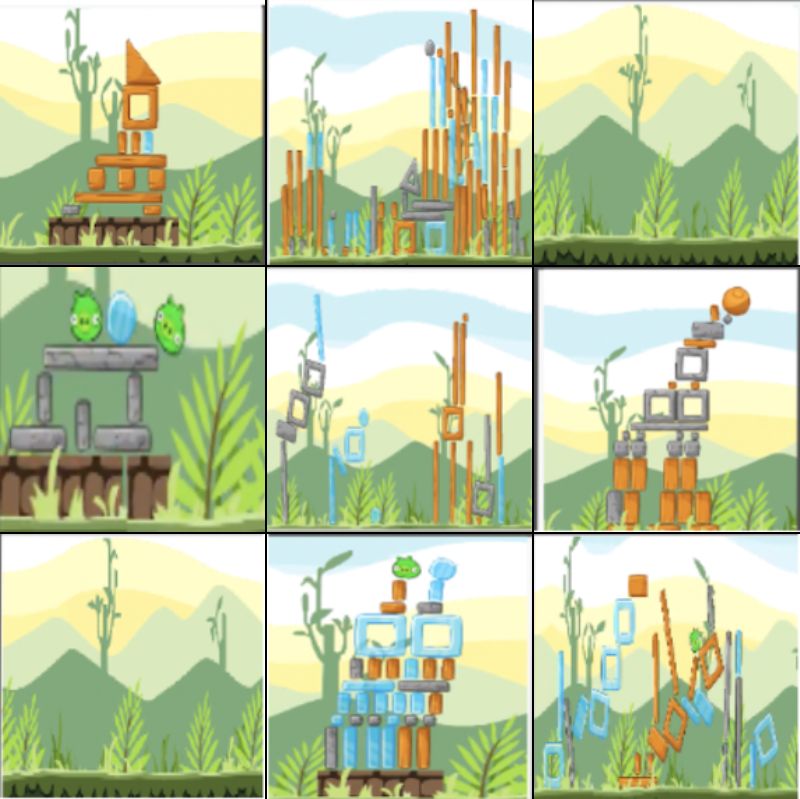}%
    \caption{Char(S)}%
    \label{fig:chars_l}%
  \end{subfigure}%
  \begin{subfigure}{0.33\hsize}%
    \centering%
    \includegraphics[width=\hsize]{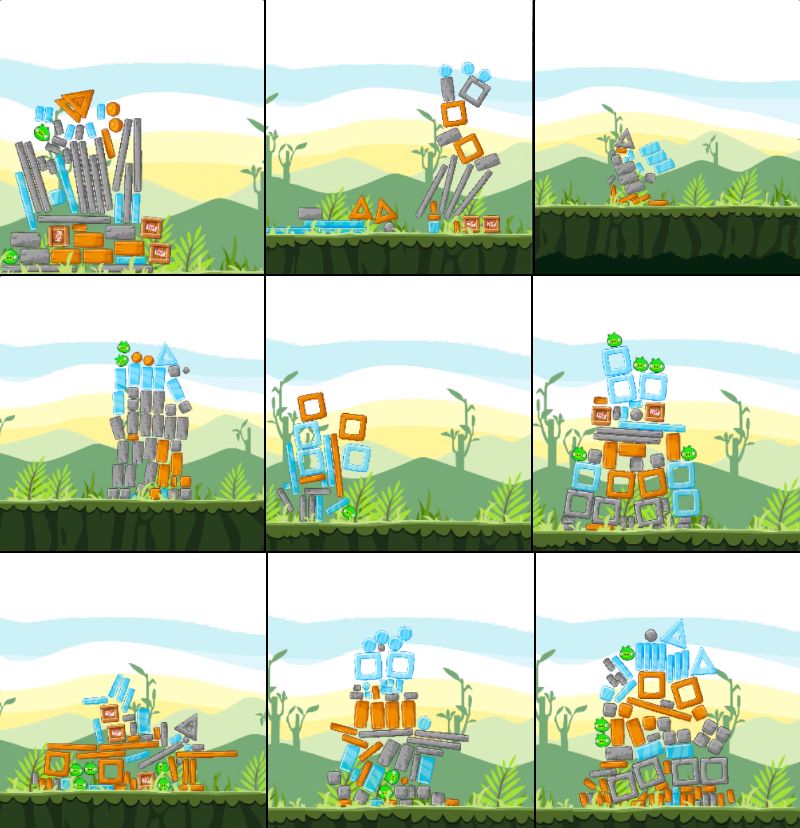}%
    \caption{Image(D)}%
    \label{fig:image_d_l}%
  \end{subfigure}
  \begin{subfigure}{0.33\hsize}%
    \centering%
    \includegraphics[width=\hsize]{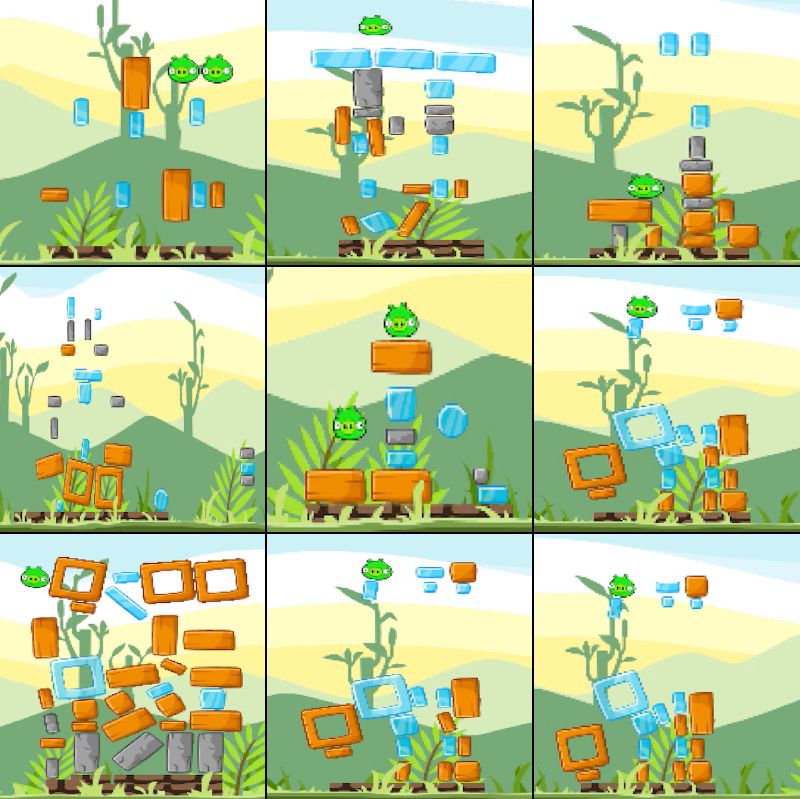}%
    \caption{Image(S)}%
    \label{fig:image_s_l}%
  \end{subfigure}
  \caption{Levels generated by each generator trained on a large dataset.}
  \label{fig:enc-all-level-l}
\end{figure*}

In this section, we compare the performance of each generator when trained on a large dataset.
In an experiment to compare the diversity of the levels generated by each generator, we generated 1000 levels for each generator and compared the number of unigrams and bigrams. Further, we randomly selected 1000 of these levels from the training dataset and counted the number of unigrams and bigrams in the selected levels.
In addition, we generated 60 of these levels using each of our generators, and examined the number of stable levels generated.
The results of each experiment are summarized in \Cref{tab:large-gram-num} and \Cref{tab:large-stable-rate}. The levels generated by each generator are shown in \Cref{fig:enc-all-level-l}.
\begin{table}[t]
  \begin{center}
    \caption{Numbers of uni-gram and bi-gram types present in the vector output from each generator.}
    \begin{tabular}{cccccc} \toprule
      & Training & \bf{Proposed} & W/o WE & Char(D) & Char(S) \\ \midrule
      Uni-gram & 6202 & 5893  & 6075 & 7335 & 4992\\
      Bi-gram & 7975 & 7470  & 7693 & 10031 & 6180\\ \bottomrule
    \end{tabular}
    \label{tab:large-gram-num}
  \end{center}
\end{table}
\begin{table}[t]
  \begin{center}
    \caption{Numbers of stable levels among $60$ levels generated from each generator.
      We consider a level to be stable if it does not fall down after the game starts.}
    \begin{tabular}{cccccc} \toprule
       \bf{Proposed} & W/o WE & Char(D) & Char(S) & Image(D) & Image(S)      \\ \midrule
          \bf{57}   & 22 & 24 & 18 & 4 & 7           \\ \bottomrule
    \end{tabular}
    \label{tab:large-stable-rate}
  \end{center}
\end{table}

\begin{figure}[t]
  \begin{subfigure}{\hsize}%
    \centering%
    \includegraphics[scale=0.5]{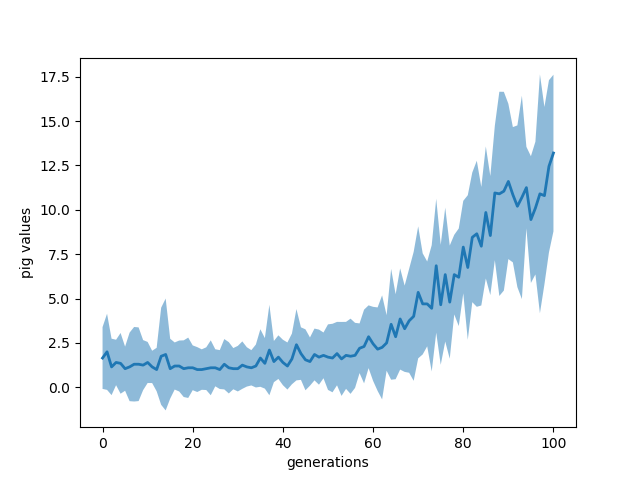}%
    \caption{Number of pig blocks}%
    \label{fig:lvepig_l}%
  \end{subfigure}%
  \\
  \begin{subfigure}{\hsize}%
    \centering%
    \includegraphics[scale=0.5]{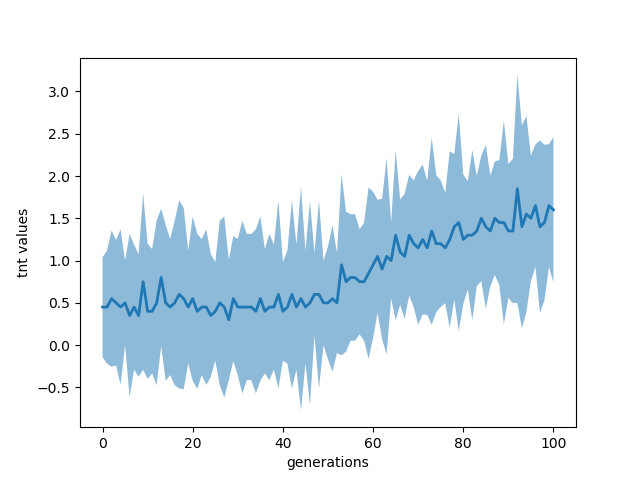}%
    \caption{Number of TNT blocks}%
    \label{fig:lvetnt_l}%
  \end{subfigure}
  \caption{Results of LVE(not simulated). Average (solid line) and standard deviation (band) of the corresponding numbers in the $10$ levels generated using the best solution for each generation.}
\end{figure}

\begin{figure}[t]
  \begin{center} 
    \centerline{\includegraphics[scale=0.3]{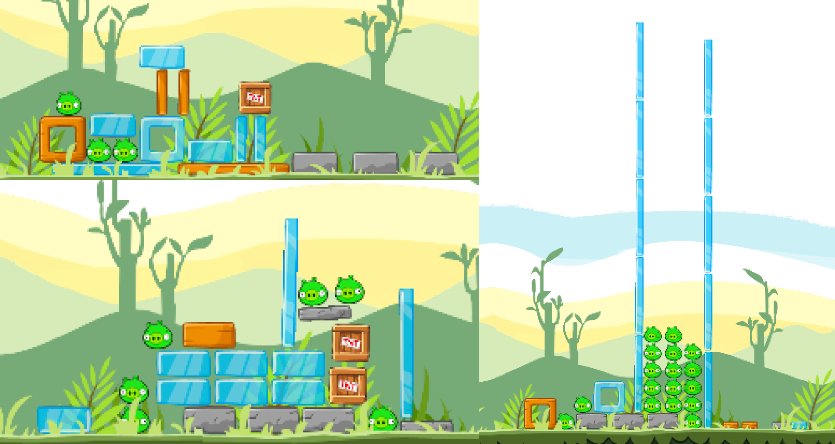}}
    \caption{Results of maximization of the number of pig blocks. Levels generated at generation 60 (top left), 80 (lower left), and generation 100 (right) are displayed.}
    \label{fig:lvepiglevel_l}
  \end{center}
\end{figure}

Further, we experimented with a non-simulation LVE using the proposed generator trained on a large dataset.
The results are shown in \Cref{fig:lvepig_l} and \Cref{fig:lvetnt_l}; the level generated during the maximization of the number of pigs is shown in \Cref{fig:lvepiglevel_l}.

\begin{table}[t]
  \begin{center}
    \caption{Setting up VAE learning for small datasets and large datasets.}
  \begin{tabular}{|c|c|c|}
  \hline
                     & Small dataset & Large dataset \\ \hline
  Dataset size       & 180           & 10000         \\ \hline
  Vocabulary size    & 1503          & 54862         \\ \hline
  Epoch              & 500           & 100           \\ \hline
  KL loss ignored    & 250           & 50            \\ \hline
  Latent vector dim  & 60            & 300           \\ \hline
  Word embedding dim & 50            & 300           \\ \hline
  Word drop rate     & 0.3           & 0.5           \\ \hline
  Batch size         & 20            & 10            \\ \hline
  \end{tabular}
  \label{tab:vaesetting}
\end{center}
\end{table}

\section{Experimental setting details}\label{apdx:setting}
\Cref{tab:vaesetting} summarizes the hyperparameter values used in our experiments.

\end{document}